\documentclass[journal]{IEEEtran}
\usepackage{bm}

\usepackage{amsfonts}
\usepackage{graphicx}
\usepackage{makecell}
\usepackage{tikz}
\usepackage{collcell}
\usepackage{colortbl}
\usepackage{pgfplots}
\usepackage{pgfplotstable}
\usepackage{multirow}
\usepackage{url}
\usepackage[all]{xy}
\usepackage{cite}
\usepackage{booktabs}
\usepackage{adjustbox}
\usepackage{tikz}
\usepackage{arydshln}
\usepackage{epstopdf}
\usetikzlibrary{arrows,positioning}
\usetikzlibrary{plotmarks}
\usetikzlibrary{calc}

\usetikzlibrary{patterns}
\usetikzlibrary{shapes.geometric}
\pgfdeclarelayer{bg}    
\pgfsetlayers{bg,main}

\newcommand{\WindowLength}{\ell}

\newcommand{\term}[1]{\mathfrak{#1}}
\newcommand{\zbior}[1]{\mathbb{#1}}  
\newcommand{\macierz}[1]{{\boldsymbol{\mathrm{#1}}}}
\newcommand{\wektor}[1]{\macierz{\MakeLowercase{#1}}}
\newcommand{\newItem}[2]{%
  \expandafter\def\csname #1\endcsname {\MakeLowercase{#2}} %
  \expandafter\def\csname n#1\endcsname {\MakeUppercase{#2}} %
  \expandafter\def\csname zb#1\endcsname {\zbior{\MakeUppercase{#2}}} %
  \expandafter\def\csname t#1\endcsname {\term{\MakeLowercase{#2}}} %
  \expandafter\def\csname m#1\endcsname {\macierz{\MakeUppercase{#2}}} %
  \expandafter\def\csname w#1\endcsname {\wektor{#2}} %
  }
  \newItem{Atrybut}{d_f}
\newItem{Deskryptor}{x}
\newItem{Regula}{l}
\newItem{Przeslanka}{a}
\newItem{Konsekwencja}{b}
\newItem{Implikacja}{b'}
\newItem{Zespolona}{c}
\newItem{Rzeczywista}{r}
\newItem{Klasa}{\delta}
\newItem{Decyzja}{y}
\newItem{Przyklad}{a}
\newItem{Test}{a}
\newItem{Klaster}{c}
\newItem{Rozmycie}{s}
\newItem{Centre}{v}
\newItem{Membership}{u}
\newItem{ClusterMembership}{\mu}
\newItem{Weight}{z}
\newItem{Krotka}{x}
\newItem{Width}{w}
\newItem{Wspolczynnik}{p}
\newItem{Obiekt}{k}
\newItem{Odleglosc}{t}
\newItem{Parameter}{p}

\newcommand{\TrainingSet}{\bm{T}} 

\newcommand{\ValidationSet}{\bm{V}} 

\newcommand{\TestSet}{\Psi} 

\ifCLASSINFOpdf
\else
\fi

\pgfplotstableset{
    /color cells/min/.initial=0,
    /color cells/max/.initial=1000,
    /color cells/textcolor/.initial=,
    color cells/.code={%
        \pgfqkeys{/color cells}{#1}%
        \pgfkeysalso{%
            postproc cell content/.code={%
                \begingroup
                \pgfkeysgetvalue{/pgfplots/table/@preprocessed cell content}\value
\ifx\value\empty
\endgroup
\else
                \pgfmathfloatparsenumber{\value}%
                \pgfmathfloattofixed{\pgfmathresult}%
                \let\value=\pgfmathresult
                \pgfplotscolormapaccess
                    [\pgfkeysvalueof{/color cells/min}:\pgfkeysvalueof{/color cells/max}]%
                    {\value}%
                    {\pgfkeysvalueof{/pgfplots/colormap name}}%
                \pgfkeysgetvalue{/pgfplots/table/@cell content}\typesetvalue
                \pgfkeysgetvalue{/color cells/textcolor}\textcolorvalue
                \toks0=\expandafter{\typesetvalue}%
                \xdef\temp{%
                    \noexpand\pgfkeysalso{%
                        @cell content={%
                            \noexpand\cellcolor[rgb]{\pgfmathresult}%
                            \noexpand\definecolor{mapped color}{rgb}{\pgfmathresult}%
                            \ifx\textcolorvalue\empty
                            \else
                                \noexpand\color{\textcolorvalue}%
                            \fi
                            \the\toks0 %
                        }%
                    }%
                }%
                \endgroup
                \temp
\fi
            }%
        }%
    }
}

\begin{document}
\title{Transfer Learning for Segmenting Dimensionally-Reduced Hyperspectral Images}
\author{Jakub Nalepa,~\IEEEmembership{Member,~IEEE}, 
        Michal Myller,
        and Michal Kawulok,~\IEEEmembership{Member,~IEEE}
\thanks{This work was funded by European Space Agency (HYPERNET project).}
\thanks{J.~Nalepa, M.~Myller, and M.~Kawulok are with Silesian University of Technology, Gliwice, Poland (e-mail: \{jnalepa, michal.kawulok\}@ieee.org), and KP Labs, Gliwice, Poland (\{jnalepa, mmyller, mkawulok\}@kplabs.pl).}
}


\maketitle
\begin{abstract}
Deep learning has established the state of the art in multiple fields, including hyperspectral image analysis. However, training large-capacity learners to segment such imagery requires representative training sets. Acquiring such data is human-dependent and time-consuming, especially in Earth observation scenarios, where the hyperspectral data transfer is very costly and time-constrained. In this letter, we show how to effectively deal with a limited number and size of available hyperspectral ground-truth sets, and apply transfer learning for building deep feature extractors. Also, we exploit spectral dimensionality reduction to make our technique applicable over hyperspectral data acquired using different sensors, which may capture different numbers of hyperspectral bands. The experiments, performed over several benchmarks and backed up with statistical tests, indicated that our approach allows us to effectively train well-generalizing deep convolutional neural nets even using significantly reduced data.
\end{abstract}


\IEEEpeerreviewmaketitle

\section{Introduction} \label{sec:intro}

Hyperspectral imaging (HSI) has been continuously gaining research attention due to the amount of information it conveys. Also, remote sensors are being developed at an enormous speed, and acquisition of HSI, with up to hundreds of spectral bands over a given spatial area, is much more affordable nowadays. \emph{Classification} and \emph{segmentation}\footnote{By \emph{classification} we mean assigning a class label to a specific HSI pixel, whereas by \emph{segmentation}---finding the boundaries of objects belonging to different classes in HSI. Hence, segmentation involves classification.} of HSI help us understand the underlying materials, and can be exploited in multiple fields including chemistry, biology, medicine, document imaging, food quality control, and many more~\cite{8314827}. In Earth observation applications, HSI can provide extremely detailed information on the Earth peculiarities, and may be utilized in an array of use cases, encompassing precision agriculture, managing environmental disasters, tracking volcano activities, military defense applications, and many more.

HSI classification and segmentation algorithms can be divided into conventional machine learning~\cite{Bilgin2011,Dundar2018}, and deep learning-powered techniques~\cite{Chen2015,Zhao2016,Zhong2017,Mou2017,Santara2017,Lee2017,Gao_2018}, with the latter constituting the current research mainstream. Deep learning has established the state of the art in a variety of fields, consistently outperforming techniques which use hand-crafted features. However, to effectively deploy such deep models in practice, we need large and representative ground-truth training sets. It is a significant obstacle in hyperspectral Earth observation analysis, where transferring hyperspectral data back to Earth is extremely costly. A problem of efficient hyperspectral data volume reduction (to enable its feasible storage and transmission from a satellite) can be tackled from multiple angles, e.g.,~by reduction of digital precision from native, usually 14- or 12-bit, down to 8- or even 1-bit, elimination of low-variance components using principal component analysis, or reduction of spectral resolution by band selection~\cite{DBLP:journals/corr/abs-1811-02667}. Annotating HSI by humans is error-prone and requires building a full understanding of the materials presented in a scanned region, therefore involves acquiring observational ground-sensor data. These difficulties are reflected in a very limited number of ground-truth hyperspectral sets---in~\cite{Nalepa2019GRSL}, we analyzed 17 recently-published HSI segmentation papers in which \emph{seven} benchmarks were exploited, and only \emph{three} of them can be considered ``widely-used'': Pavia University (utilized in 15 works), Indian Pines (8), and Salinas Valley (5).

In this letter, we tackle both problems of (i)~limited number of ground-truth hyperspectral sets, and (ii)~large volumes of such data. We employ \emph{transfer learning} to make convolutional neural networks (CNNs) applicable in supervised HSI segmentation with limited ground truth (Section~\ref{sec:method}). First, we train the feature extraction part of a CNN over a source (larger) set, and then we fine tune its classification part over the target (much smaller) set. Since different sensors acquire HSI with different spectral characteristics, we exploit our recent algorithm for simulating multispectral image (MSI) from its hyperspectral counterpart~\cite{Marcinkiewicz2019IGARSS}, and reduce the dimensionality of both source and target sets to the same number of bands. This operation allows us to build extractors that are applicable to \emph{any} HSI, once this HSI is reduced to the assumed number of bands. Also, it brings the possibilities of on-board data reduction executed before transferring the acquired data from an imaging satellite. Although there exist works which show the usefulness of transfer learning in HSI segmentation in various fields~\cite{Venkatesan2019}, they are focused on applying this technique to different deep architectures~\cite{8245897,DBLP:journals/corr/abs-1901-08658,8547139,DBLP:journals/corr/abs-1904-02454,JARS2019_transfer}. To the best of our knowledge, our approach is the first which comprehensively combines effective HSI data reduction and transfer learning. The experiments showed that the proposed algorithm leads to well-generalizing convolutional models, and the HSI reduction does not adversely affect their performance (Section~\ref{sec:experiments}).

\section{Transfer Learning for HSI Segmentation}\label{sec:method}

\begin{figure}[ht!]
\centering
\hspace*{-1cm}
\resizebox{1.1\columnwidth}{!}{

\newcommand{\largefig}{120pt}
\newcommand{\mediumfig}{60pt}
\newcommand{\inputfig}{35pt}

\newcommand{\resnettext}{Reconstruction with ResNet}
\newcommand{\registrationtext}{Image registration}
\newcommand{\shifttext}{Shifts $(x,y)$}
\newcommand{\evoimtext}{EvoIM iterative image filtering}
\newcommand{\groundtruthtext}{$M$ ground-truth high-resolution images $\mathcal{I}^{(h)}$}

\newcommand{\blocksep}{45pt}
\newcommand{\textsep}{15pt}
\newcommand{\linesep}{15pt}

\begin{tikzpicture}[scale=0.9,
        image/.style={inner sep=0pt,draw=white,very thick},
        mytext/.style={inner sep=1pt,draw=none},
        mygradient/.style={left color=white, right color=black, shading angle=0, draw=red},
        whitefont/.style={text=white,font=\footnotesize},
        artefact/.style={rectangle,draw=black,fill=gray!20,inner sep=5pt,minimum height=32pt,minimum width=40pt,text width=50pt,text badly centered,font=\normalsize,thick},
        output/.style={rectangle,draw=black,fill=gray!20,inner sep=5pt,minimum height=32pt,minimum width=40pt,text width=50pt,text badly centered,font=\normalsize,thick},
        ga/.style={rectangle,draw=black,fill=red!20,inner sep=5pt,minimum height=32pt,minimum width=40pt,text width=100pt,text badly centered,font=\normalsize,thick,rounded
        corners=16pt},
        algstep/.style={rectangle,draw=black,fill=red!20,inner sep=5pt,minimum height=27pt,minimum width=40pt,text width=60pt,text badly centered,font=\normalsize,thick,rounded
        corners=8pt},
        plain/.style={rectangle,text width=40pt,text badly centered,font=\small},
        myarrow/.style={thick}]

\node (train_extractor) [artefact] {Train extractor};
\node (train_data_A) [algstep, below=of train_extractor] {Source training data};

\node (train_classifier) [artefact, right=of train_extractor] {Fine tune classifier};
\node (train_data_B) [algstep, below=of train_classifier] {Target training data};

\node (test) [artefact, right=of train_classifier] {Classify unseen data};
\node (test_data) [algstep, below=of test] {Test data};

\node (predicted_label) [algstep, right=of test] {Predicted output};

\draw [myarrow,->] (train_extractor) -- (train_classifier);
\draw [myarrow,->] (train_classifier) -- (test);
\draw [myarrow,->,dashed] (test) -- (predicted_label);

\draw [myarrow,->,dashed] (test_data) -- (test);
\draw [myarrow,->,dashed] (train_data_B) -- (train_classifier);
\draw [myarrow,->,dashed] (train_data_A) -- (train_extractor);

\end{tikzpicture}
}
\caption{In our transfer-learning technique, we train the feature extractor over a source training data, and fine tune the classifier over the target data. The spectral dimensionality (i.e.,~the number of bands) of both sets is the same. Note that a different number of classes in the source and target sets is not an issue, since we fine tune the CNN classification part over the target data.}\label{fig:transfer_learning}
\end{figure}
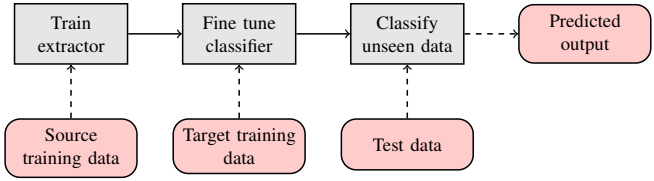

Transfer learning helps us build large-capacity learners, e.g.,~deep neural networks, over \emph{small} training data. In our approach (Fig.~\ref{fig:transfer_learning}), we train the deep feature extractor over a source hyperspectral training data (containing $t_S$ training examples), and fine tune the classification part of a CNN over the target training data ($t_T$ examples, where $t_S\gg t_T$). The fine-tuned CNN is used to classify the incoming test examples.

Since different hyperspectral sensors acquire HSI with different spectral characteristics, the number of bands in a source training data (denoted as $b_S$) is likely to be different than the number of bands in the target set ($b_T$). To make our method applicable to \emph{any} HSI, we simulate the MSI data based on the original hyperspectral imagery by using non-overlapping sliding windows of a size $\WindowLength$~\cite{Marcinkiewicz2019IGARSS}, and reduce the number of bands in both source and target sets to $b_M$. Let us consider a source HSI with $b_S$ bands $b_i$, $i=0,1,\dots,b_S-1$. Its simulated multispectral counterpart will contain $b_M$ bands, where $b_M=\lceil b_S/\WindowLength\rceil$. The corresponding multispectral bands become $b_i'=f(b_{i\cdot\WindowLength}, b_{i\cdot\WindowLength+1}, ... , b_{i\cdot\WindowLength+\WindowLength - 1})$, where $f$ is a function which transforms $\WindowLength$ consecutive HSI bands into simulated ones. Although $f$ may be conveniently updated to any function that maps the neighboring signals into an aggregated signal~\cite{Marcinkiewicz2019IGARSS}, we perform the averaging across $\WindowLength$ bands in a window. It can be seen as having wide bands covering the spectrum instead of more narrower bands. Sensors which are sensitive at broader range of wavelengths gather more light, thus they can increase the signal-to-noise ratio. Usually, the sensor sensitivity is wavelength-dependent---averaging the neighboring bands can be interpreted as an approximation of wider bands that could have been acquired by a multispectral sensor.

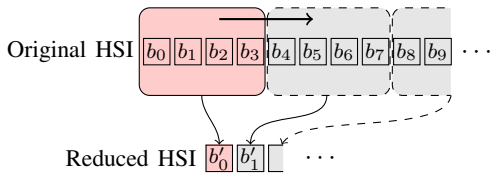
\begin{figure}[ht!]

\newcommand{\dd}{0.8}
\newcommand{\ddx}{0.02}
\newcommand{\ws}{0.085}
\newcommand{\shiftx}{0.1}
\newcommand{\shifty}{0}

\begin{tikzpicture}[scale=0.5,
        band/.style={draw=black,inner sep=0pt,minimum height=10pt,minimum width=10pt,trapezium left angle=120, trapezium right angle=60,font=\footnotesize},
        mytext/.style={text width=70pt,align=right},
        ]

    \node (t1) [band] {$b_0$};
    \node (t2) at ($(t1.north east) + (\shiftx,\shifty) $) [band, anchor=north west] {$b_1$};
    \node (t3) at ($(t2.north east) + (\shiftx,\shifty) $) [band, anchor=north west] {$b_2$};
    \node (t4) at ($(t3.north east) + (\shiftx,\shifty) $) [band, anchor=north west] {$b_3$};
    \node (t5) at ($(t4.north east) + (\shiftx,\shifty) $) [band, anchor=north west] {$b_4$};
    \node (t6) at ($(t5.north east) + (\shiftx,\shifty) $) [band, anchor=north west] {$b_5$};
    \node (t7) at ($(t6.north east) + (\shiftx,\shifty) $) [band, anchor=north west] {$b_6$};
    \node (t8) at ($(t7.north east) + (\shiftx,\shifty) $) [band, anchor=north west] {$b_7$};
    \node (t9) at ($(t8.north east) + (\shiftx,\shifty) $) [band, anchor=north west] {$b_8$};
    \node (t10) at ($(t9.north east) + (\shiftx,\shifty) $) [band, anchor=north west] {$b_9$};

    \node [mytext] at (t1.west) [anchor=east] {\small Original HSI};
    \node at (t10.east) [anchor=west] {$\cdots$};

\begin{pgfonlayer}{bg}
    \draw [color=black,fill=red!20, rounded corners=4pt]  ($(t1.north west) + (-0.1,\dd)$) rectangle ($(t4.south east) + (\ddx,-\dd)$);
    \draw [color=black,fill=gray!20,dashed, rounded corners=4pt]  ($(t5.north west) + (-\ddx,\dd)$) rectangle ($(t8.south east) + (\ddx,-\dd)$);
    \draw [color=black,fill=gray!20,dashed, rounded corners=4pt]  ($(t9.north west) + (-\ddx,\dd)$) rectangle ($(t10.south west) + (1,-\dd)$);
    \draw [color=white,fill=white, thick] ($(t10.north east) + (0,\dd)$) rectangle ($(t10.south east) + (1,-\dd)$);
\end{pgfonlayer}
    \draw [->,thick] ($(t3.north) + (0,0.5)$) -> ($(t6.north) + (0,0.5)$);

    \node (b1) [band, below=30pt of t3] {$b'_0$};
    \node (b2) at ($(b1.north east) + (\shiftx,\shifty) $) [band, anchor=north west] {$b'_1$};
    \node (b3) at ($(b2.north east) + (\shiftx,\shifty) $) [band, anchor=north west] {};

    \draw[->] ($(t2.south east) + (0,-\dd)$)  .. controls +(-90:0.5) and +(90:0.5) .. (b1.north);
    \draw[->] ($(t6.south east) + (0,-\dd)$)  .. controls +(-90:0.9) and +(90:1.0) .. (b2.north);
    \draw[->,dashed] ($(t10.south east) + (-0.02,-\dd)$)  .. controls +(-100:1.1) and +(60:1.0) .. (b3.north);

\begin{pgfonlayer}{bg}
    \draw [fill=red!20,draw=none]  (b1.north west) rectangle (b1.south east);
    \draw [fill=gray!20,draw=none]  (b2.north west) rectangle (b2.south east);
    \draw [fill=gray!20,draw=none]  (b3.north west) rectangle (b3.south east);
\end{pgfonlayer}
    \draw [color=white,fill=white, thick] ($(b3.north east) + (-0.3,0)$) rectangle ($(b3.south east) + (1,0)$);
    \node at (b3.east) [anchor=west] {$\cdots$};
    \node [mytext] at (b1.west) [anchor=east] {\small Reduced HSI};

\end{tikzpicture}
\caption{Reducing the dimensionality of an input hyperspectral data by simulating its multispectral version using our sliding-window approach ($\WindowLength=4$).}\label{fig:simulation_example}
\end{figure}

In Fig.~\ref{fig:simulation_example}, we present an example process of simulating MSI from its hyperspectral counterpart using the proposed sliding-window approach (with $\WindowLength=4$). The reduction ratio is dependent on the window size $\WindowLength$, and increasing $\WindowLength$ will lead to a lower number of simulated bands. This reduction is crucial in Earth observation scenarios to effectively transfer the acquired HSI from a satellite. Let us assume we want to capture a $2048\times 2048$ 12-bit HSI with 200 bands, and send it back to Earth. This would give $2048\cdot 2048 \cdot 200 \cdot 12\approx 10$ gigabits for transmission. If we could use an X-band link with 3 Mbps nominal downlink speed, it would require 3355 s (56 min) for a single scene. Simulating MSI with 20 bands (the volume is reduced $10\times$), ideally without affecting the performance of a segmentation algorithm applied over this data, would greatly decrease this time and make it much more affordable.

\section{Experiments}\label{sec:experiments}

\begin{figure}[ht!]

    \newcommand{\dd}{0.8}
\newcommand{\ddx}{0.02}
\newcommand{\ws}{0.085}
\newcommand{\shiftx}{0.1}
\newcommand{\shifty}{0}

        \centering
        \resizebox{0.84\columnwidth}{!}{%
            \begin{tikzpicture}[scale=0.8,
                    input/.style={rectangle,draw=black,fill=red!20,inner sep=5pt,minimum height=50pt,minimum width=10pt,text width=5pt,text badly centered,thick},
                    output/.style={rectangle,draw=black,fill=green!20,inner sep=5pt,minimum height=50pt,minimum width=10pt,text width=5pt,text badly centered,thick},
                    convolution/.style={rectangle,draw=black,fill=orange!20,inner sep=5pt,minimum height=50pt,minimum width=10pt,text width=5pt,text badly centered,thick},
                    maxpooling/.style={rectangle,draw=black,fill=cyan!20,inner sep=5pt,minimum height=50pt,minimum width=10pt,text width=5pt,text badly centered,thick},
                    batchNorm/.style={rectangle,draw=black,fill=blue!20,inner sep=5pt,minimum height=50pt,minimum width=10pt,text width=5pt,text badly centered,thick},
                    fullyconnected/.style={rectangle,draw=black,fill=gray!20,inner sep=5pt,minimum height=50pt,minimum width=10pt,text width=5pt,text badly centered,thick},
                    softmax/.style={rectangle,draw=black,fill=gray!20,inner sep=5pt,minimum height=50pt,minimum width=10pt,text width=5pt,text badly centered,thick},
                    myarrow/.style={thick},
                    dottedarrow/.style={dotted, thick}]

                \newcommand{\HSIpixel}{{\rotatebox{90}{{HSI pixel}}}}
                \newcommand{\class}{{\rotatebox{90}{{Class}}}}
                \newcommand{\sep}{20}
                \newcommand{\bigSep}{20}
                \newcommand{\convolution}{{\rotatebox{90}{{Conv}}}}
                \newcommand{\maxpooling}{{\rotatebox{90}{{Max pool}}}}
                \newcommand{\batchnormalization}{{\rotatebox{90}{{BN}}}}
                \newcommand{\deconvolution}{{\rotatebox{90}{{Deconv}}}}
                \newcommand{\fullyconnected}{{\rotatebox{90}{{FC}}}}
                \newcommand{\softmax}{{\rotatebox{90}{{Softmax}}}}
                \newcommand{\convDescrSep}{1}
                \mathchardef\mhyphen="2D 

                \node (in) [input] {$\HSIpixel$};
                \node (conv1) [convolution, right=30 pt of in]{\convolution};
                \node (convDescription) at ($(conv1.south) - (0 pt, \convDescrSep pt)$)[anchor=north] {{\tiny$\bm{s=1\times 1\times 5}$}};
                \node (convDescription2) at ($(convDescription.south) + (0 pt, 5 pt)$)[anchor=north] {{\tiny$\bm{n=200}$}};
                \node (batchnorm1) [batchNorm, right=\sep pt of conv1]{\batchnormalization};
                \node (maxpooling1) [maxpooling, right=\sep pt of batchnorm1]{\maxpooling};
                \node (convDescription3) at ($(maxpooling1.south) - (0 pt, \convDescrSep pt)$)[anchor=north] {{\tiny$\bm{s=2\times 2}$}};
                \node (fc1) [fullyconnected, right=\sep pt of maxpooling1]{\fullyconnected};
                \node (convDescription3) at ($(fc1.south) - (0 pt, \convDescrSep pt)$)[anchor=north] {{\tiny$\bm{l_1=512}$}};
                \node (convDescription4) at ($(convDescription3.south) + (0 pt, 5 pt)$)[anchor=north] {{\tiny$\bm{l_2=128}$}};
                \node (softmax1) [softmax, right=\sep pt of fc1]{\softmax};
                \node (out) [output, right=\sep pt of softmax1]{\class};
                \node (blockDescription) at ($(batchnorm1.north) + (0 pt, 20 pt)$)[anchor=north] {{\textbf{Building block}}};

                \begin{pgfonlayer}{bg}
                \draw [color=black,fill=yellow!20, rounded corners=4pt]  ($(conv1.north west) + (-0.7,\dd)$) rectangle ($(maxpooling1.south east) + (0.5,-\dd)$);
                \end{pgfonlayer}

                \draw[->,myarrow] (in) -- (conv1);
                \draw[->,myarrow] (conv1) -- (batchnorm1);
                \draw[->,myarrow] (batchnorm1) -- (maxpooling1);
                \draw[->,myarrow] (maxpooling1) -- (fc1);
                \draw[->,myarrow] (fc1) -- (softmax1);
                \draw[->,myarrow] (softmax1) -- (out);
            \end{tikzpicture}
        } \vspace*{-0.2cm}\caption{1D-CNN with $n$ kernels in the convolutional layer ($s$ stride) and $l_1$ and $l_2$ neurons in the fully-connected (FC) layers. BN is batch normalization.} \label{fig:1d_network}
    \end{figure}
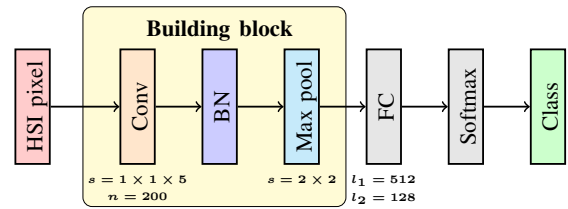

The main objective of our experiments was to verify if transfer learning applied over reduced HSI can be effectively used to get well-generalizing CNNs. We investigated two CNNs: our 1D-CNN (Fig.~\ref{fig:1d_network}) with varying numbers of building blocks (one, two, and three) constituting the feature extractor and two fully-connected (FC) layers followed by softmax in the classification part, alongside a state-of-the-art CNN (we call it PT-CNN) with one, two, and three convolutional layers acting as a feature extractor (for simplicity, we refer to these layers as building blocks too), followed by three FC layers and softmax, which was applied to hyperspectral transfer learning in~\cite{8245897}. Therefore, the main difference between 1D-CNN and PT-CNN is the lack of pooling layers in the latter network. Larger CNNs are unlikely to be deployed in hardware-constrained on-board settings due to their memory requirements. The deep nets were implemented in \texttt{Python 3.6}, and the training (ADAM, learning rate of $10^{-4}$, $\beta_1 = 0.9$, $\beta_2 = 0.999$) stops, if after 25 epochs the overall accuracy over the validation set (random subset of the training set) does not change. The experiments were run on NVIDIA GeForce GTX~1060.

\begin{table*}[ht!]
\newcommand{\mybackground}{35}
	\scriptsize
	\centering
	\caption{The results obtained for all configurations of our 1D-CNN architecture.}
	\label{tab:summary_our_network}
	\renewcommand{\tabcolsep}{0.057cm}
	\begin{tabular}{r|r|r|r|r|r|r|r|r|r|r|r|r|r|r|r|r|r|r|r|r|r|r|r|r|r|r|r}
\hline
CNN$\rightarrow$ & \multicolumn{9}{c|}{1 block} & \multicolumn{9}{c|}{2 blocks} & \multicolumn{9}{c}{3 blocks}\\
\hline
 Set$\rightarrow$	& \multicolumn{3}{c|}{Sa}		& \multicolumn{3}{c|}{PU} & \multicolumn{3}{c|}{IP} &
\multicolumn{3}{c|}{Sa}		& \multicolumn{3}{c|}{PU} & \multicolumn{3}{c|}{IP} &
\multicolumn{3}{c|}{Sa}		& \multicolumn{3}{c|}{PU} & \multicolumn{3}{c}{IP} \\	
\hline

&			OA		&AA	 & $\kappa$	&OA 	&AA	& $\kappa$	 &OA		&AA	& $\kappa$	
&			OA		&AA	 & $\kappa$	&OA 	&AA	& $\kappa$	 &OA		&AA	& $\kappa$	
&			OA		&AA	 & $\kappa$	&OA 	&AA	& $\kappa$	 &OA		&AA	& $\kappa$	\\
\hline
Var.$\downarrow$ &\multicolumn{27}{c}{Full spectrum}\\
\hline
B(E) &95.47&97.98&0.94&96.72&95.79&0.95&88.50&87.34&0.87&95.55&97.96&0.95&96.24&94.92&0.94&89.13&87.35&0.88&95.49&97.91&0.94&96.37&95.17&0.94&89.56&88.68&0.88\\
\hdashline
B    &\cellcolor{gray!\mybackground}89.01&\cellcolor{gray!\mybackground}94.84&\cellcolor{gray!\mybackground}0.88&\cellcolor{gray!\mybackground}89.32&92.19&0.87&76.47&\cellcolor{gray!\mybackground}81.38&0.73&\cellcolor{gray!\mybackground}90.24&\cellcolor{gray!\mybackground}95.53&\cellcolor{gray!\mybackground}0.90&90.56&92.27&\textbf{0.88}&\cellcolor{gray!\mybackground}78.45&\cellcolor{gray!\mybackground}82.03&\cellcolor{gray!\mybackground}0.75&89.64&95.51&\cellcolor{gray!\mybackground}0.89&89.66&92.09&0.87&79.33&83.71&0.76\\
\hline
 &\multicolumn{27}{c}{100 bands}\\
\hline
B(E) &95.55 & 97.97 & 0.95 & 95.32 & 93.89 & 0.86 & 90.42 & 89.74 & 0.92 & 94.37 & 97.34 & 0.94 & 95.29 & 93.24 & 0.93 & 84.61 & 81.79 & 0.85 & 93.63 & 96.75 & 0.93 & 95.28 & 93.37 & 0.94 & 81.17 & 79.44 & 0.81 \\
\hdashline
B &90.81 & 95.80 & \cellcolor{red!30}\textbf{0.91} & \cellcolor{red!30}91.02 & \cellcolor{red!30}93.06 & \cellcolor{gray!\mybackground}0.85 & 81.06 & 85.42 & 0.77 & 91.11 & 96.15 & \cellcolor{gray!\mybackground}0.90 & \cellcolor{red!30}91.03 & \cellcolor{red!30}92.73 & \cellcolor{red!30}0.87 & 81.68 & 85.97 & 0.80 & 90.04 & 95.76 & \cellcolor{gray!\mybackground}0.89 & \cellcolor{red!30}91.38 & \cellcolor{red!30}92.42 & \cellcolor{red!30}0.87 & 82.57 & 86.17 & \cellcolor{red!30}0.82 \\
Ex(Sa)   &--- & --- & --- & 90.75 & 92.35 & \cellcolor{green!40}0.86 & 81.98 & 86.82 & \cellcolor{green!40}\textbf{0.79} & --- & --- & --- & 89.43 & \cellcolor{gray!\mybackground}91.37 & 0.85 & \cellcolor{green!40}83.29 & 87.22 & \cellcolor{green!40}0.81 & --- & --- & --- & 88.29 & \cellcolor{gray!\mybackground}90.70 & 0.84 & \cellcolor{green!40}82.86 & 86.61 & 0.80 \\
Ex(PU)   &91.26 & 96.40 & 0.90 & --- & --- & --- & \cellcolor{green!40}82.29 & \cellcolor{green!40}87.25 & \cellcolor{green!40}\textbf{0.79} & 91.08 & 96.32 & \cellcolor{gray!\mybackground}0.90 & --- & --- & --- & 82.71 & \cellcolor{green!40}87.76 & 0.80 & 91.36 & 96.25 & \cellcolor{green!40}0.90 & --- & --- & --- & 82.79 & \cellcolor{green!40}87.36 & 0.80 \\
Ex(IP)   &\cellcolor{green!40}\textbf{91.81} & \cellcolor{green!40}\textbf{96.63} &  \cellcolor{green!40}\textbf{0.91} & 90.92 & 92.33 & \cellcolor{gray!\mybackground}0.85 & --- & --- & --- & \cellcolor{green!40}91.36 & \cellcolor{green!40}96.39 & \cellcolor{green!40}\textbf{0.91} & 90.90 & 92.39 & 0.84 & --- & --- & --- & \cellcolor{green!40}91.56 & \cellcolor{green!40}96.35 & \cellcolor{green!40}0.90 & 90.47 & 92.05 & \cellcolor{gray!\mybackground}0.82 & --- & --- & --- \\
\hline
 &\multicolumn{27}{c}{75 bands}\\
\hline
B(E)                &96.21 & 98.29 & 0.96 & 96.95 & 95.95 & 0.94 & 90.93 & 90.84 & 0.92 & 96.53 & 98.42 & 0.96 & 96.69 & 95.49 & 0.95 & 91.84 & 91.39 & 0.92 & 95.15 & 97.71 & 0.94 & 96.40 & 95.16 & 0.94 & 89.51 & 87.13 & 0.88 \\
\hdashline
B                   &90.34 & 95.70 & \cellcolor{red!30}0.90 & 91.38 & 93.21 & \cellcolor{red!30}\textbf{0.89} & 80.21 & 85.11 & 0.75 & 90.87 & 96.18 & \cellcolor{gray!\mybackground}0.90 & 92.11 & \cellcolor{red!30}\textbf{93.43} & 0.86 & 82.70 & 87.12 & 0.80 & 89.88 & 95.67 & \cellcolor{gray!\mybackground}0.89 & 90.81 & 92.82 & \cellcolor{red!30}\textbf{0.88} & 82.26 & 86.28 & 0.80 \\
Ex(Sa)   & --- & --- & --- & 91.98 & 93.33 & 0.88 & \cellcolor{green!40}\textbf{83.05} & \cellcolor{green!40}\textbf{87.49} & \cellcolor{green!40}\textbf{0.79} & --- & --- & --- & 91.38 & 92.98 & \cellcolor{gray!\mybackground}0.83 & \cellcolor{green!40}\textbf{84.74} & \cellcolor{green!40}88.53 & \cellcolor{green!40}\textbf{0.83} & --- & --- & --- & 91.10 & 92.84 & \cellcolor{green!40}\textbf{0.88} & \cellcolor{green!40}\textbf{85.07} & \cellcolor{green!40}\textbf{89.27} & \cellcolor{green!40}\textbf{0.83} \\
Ex(PU)   &91.32 & 96.39 & \cellcolor{green!40}0.90 & --- & --- & --- & 81.41 & 86.93 & 0.78 & 91.36 & 96.32 & \cellcolor{green!40}\textbf{0.91} & --- & --- & --- & 81.51 & 87.85 & 0.80 & \cellcolor{green!40}\textbf{91.83} & 96.47 & \cellcolor{green!40}\textbf{0.91} & --- & --- & --- & 82.86 & 87.58 & 0.81 \\
Ex(IP)   &\cellcolor{green!40}91.53 & \cellcolor{green!40}96.44 & \cellcolor{green!40}0.90 & \cellcolor{green!40}92.34 & \cellcolor{green!40}\textbf{93.49} & \cellcolor{green!40}\textbf{0.89} & --- & --- & --- & \cellcolor{green!40}92.08 & \cellcolor{green!40}96.66 & \cellcolor{green!40}\textbf{0.91} & \cellcolor{green!40}\textbf{92.16} & 93.32 & \cellcolor{green!40}0.87 & --- & --- & --- & 91.79 & \cellcolor{green!40}\textbf{96.57} & \cellcolor{green!40}\textbf{0.91} & \cellcolor{green!40}\textbf{91.47} & \cellcolor{green!40}\textbf{92.99} & 0.84 & --- & --- & --- \\
\hline
 &\multicolumn{27}{c}{50 bands}\\
\hline
B(E) &95.87 & 98.16 & 0.95 & 96.78 & 95.60 & 0.94 & 90.31 & 89.45 & 0.90 & 95.82 & 98.06 & 0.95 & 96.57 & 95.42 & 0.94 & 91.88 & 91.04 & 0.93 & 94.56 & 97.08 & 0.94 & 95.84 & 94.43 & 0.94 & 85.93 & 81.59 & 0.85 \\
\hdashline
B &89.96 & 95.48 & \cellcolor{red!30}0.90 & 90.91 & 92.92 & \cellcolor{red!30}0.87 & 78.66 & 84.87 & 0.74 & 90.51 & 95.87 & \cellcolor{gray!\mybackground}0.90 & 91.77 & 93.10 & 0.87 & 82.67 & 86.68 & 0.80 & \cellcolor{gray!\mybackground}89.21 & \cellcolor{gray!\mybackground}94.80 & \cellcolor{gray!\mybackground}0.89 & \cellcolor{red!30}90.37 & \cellcolor{red!30}92.02 & \cellcolor{red!30}0.84 & \cellcolor{gray!\mybackground}78.81 & \cellcolor{gray!\mybackground}83.31 & \cellcolor{gray!\mybackground}0.75 \\
Ex(Sa)   & --- & --- & --- & 91.55 & 93.14 & \cellcolor{green!40}0.87 & \cellcolor{green!40}81.87 & \cellcolor{green!40}87.20 & \cellcolor{green!40}\textbf{0.79} & --- & --- & --- & 91.72 & 92.96 & 0.85 & \cellcolor{green!40}84.54 & \cellcolor{green!40}\textbf{88.75} & \cellcolor{green!40}0.82 & --- & --- & --- & 88.26 & 90.84 & \cellcolor{green!40}0.84 & 81.16 & 86.34 & \cellcolor{green!40}0.80 \\
Ex(PU)   &\cellcolor{green!40}91.28 & \cellcolor{green!40}96.26 & \cellcolor{green!40}0.90 & --- & --- & --- & 81.24 & 86.09 & 0.76 & 91.55 & 96.35 & \cellcolor{green!40}\textbf{0.91} & --- & --- & --- & 83.03 & 86.93 & 0.79 & \cellcolor{green!40}91.71 & \cellcolor{green!40}96.46 & \cellcolor{green!40}\textbf{0.91} & --- & --- & --- & \cellcolor{green!40}82.13 & \cellcolor{green!40}87.14 & \cellcolor{green!40}0.80 \\
Ex(IP)   &91.14 & 96.18 & \cellcolor{green!40}0.90 & \cellcolor{green!40}\textbf{92.54} & \cellcolor{green!40}93.34 & \cellcolor{green!40}0.87 & --- & --- & --- & \cellcolor{green!40}\textbf{92.25} & \cellcolor{green!40}\textbf{96.72} & \cellcolor{green!40}\textbf{0.91} & \cellcolor{green!40}92.01 & \cellcolor{green!40}93.14 & \cellcolor{green!40}\textbf{0.88} & --- & --- & --- & 91.29 & 96.27 & 0.90 & \cellcolor{gray!\mybackground}88.18 & 90.95 & 0.83 & --- & --- & --- \\
\hline
 &\multicolumn{27}{c}{25 bands}\\
\hline
B(E)                &94.91 & 97.68 & 0.94 & 96.44 & 95.21 & 0.94 & 88.56 & 89.67 & 0.87 & 94.93 & 97.61 & 0.94 & 96.26 & 94.89 & 0.95 & 89.30 & 88.35 & 0.88 & --- & --- & ---& --- & --- & ---& --- & --- & ---\\
\hdashline
B                   &90.05 & 95.24 & \cellcolor{red!30}0.90 & 89.69 & \cellcolor{gray!\mybackground}91.99 & \cellcolor{gray!\mybackground}0.85 & \cellcolor{gray!\mybackground}76.42 & 82.54 & \cellcolor{gray!\mybackground}0.71 & 90.35 & 95.60 & \cellcolor{gray!\mybackground}0.90 & \cellcolor{red!30}91.07 & \cellcolor{red!30}92.63 & \cellcolor{red!30}0.87 & 82.94 & 87.02 & 0.80 & --- & --- & ---& --- & --- & ---& --- & --- & ---\\
Ex(Sa)   & --- & --- & --- & 91.39 & 92.92 & \cellcolor{green!40}0.86 & \cellcolor{green!40}80.28 & \cellcolor{green!40}85.46 & \cellcolor{green!40}0.77 & --- & --- & --- & 89.56 & 91.84 & 0.85 & \cellcolor{green!40}83.85 & \cellcolor{green!40}88.22 & \cellcolor{green!40}0.81 & --- & --- & ---& --- & --- & ---& --- & --- & ---\\
Ex(PU)   &90.62 & \cellcolor{green!40}95.85 & \cellcolor{green!40}0.90 & --- & --- & --- & 79.74 & 84.29 & \cellcolor{green!40}0.77 & 91.50 & 96.20 & \cellcolor{gray!\mybackground}0.90 & --- & --- & --- & 82.68 & 87.78 & 0.77 & --- & --- & ---& --- & --- & ---& --- & --- & ---\\
Ex(IP)   &\cellcolor{green!40}90.74 & 95.79 & \cellcolor{green!40}0.90 & \cellcolor{green!40}91.76 & \cellcolor{green!40}93.00 & \cellcolor{green!40}0.86 & --- & --- & --- & \cellcolor{green!40}91.90 & \cellcolor{green!40}96.58 & \cellcolor{green!40}\textbf{0.91} & \cellcolor{gray!\mybackground}88.75 & 91.64 & \cellcolor{gray!\mybackground}0.83 & --- & --- & --- & --- & --- & ---& --- & --- & ---& --- & --- & ---\\
\hline
\multicolumn{28}{c}{\scriptsize \textbf{How to read this table:} The \emph{globally} best result (across HSI and simulated MSI), excluding B(E), in each column is boldfaced, and the background of the worst cell is grayed. }\\
\multicolumn{28}{c}{\scriptsize For \emph{each number} (100, 75, 50, and 25) of simulated multispectral bands, the background of the cell with the best result is colored---if the best result is obtained using transfer }\\
\multicolumn{28}{c}{\scriptsize learning, the background is green. If the best result is obtained using a model trained over the B division of the target set (i.e.,~\emph{without} transfer learning), the background is red.}
	\end{tabular}
\end{table*}

In this work, we exploited three most popular HSI benchmarks from the literature (see Section~\ref{sec:intro}): (i)~Salinas Valley, USA ($217\times 512$ pixels, AVIRIS sensor; $\left|\left|\TrainingSet\right|\right|=4320$, $\left|\left|\ValidationSet\right|\right|=480$, $\left|\left|\TestSet\right|\right|=49329$) presenting different sorts of vegetation (16 classes, 224 bands, 3.7 m spatial resolution); (ii)~Indian Pines, USA ($145\times 145$, AVIRIS; $\left|\left|\TrainingSet\right|\right|=2444$, $\left|\left|\ValidationSet\right|\right|=271$, $\left|\left|\TestSet\right|\right|=7534$) covering agriculture and forest (16 classes, 200 channels, 20~m); (iii)~Pavia University, Italy ($340\times 610$, ROSIS; $\left|\left|\TrainingSet\right|\right|=2025$, $\left|\left|\ValidationSet\right|\right|=225$, $\left|\left|\TestSet\right|\right|=40526$) presenting urban scenery (9 classes, 103 channels, 1.3 m). For training feature extractors, we randomly split the source set into non-overlapping training ($\TrainingSet$; balanced), validation ($\ValidationSet$), and test ($\TestSet$) sets containing 80\%, 10\%, and 10\% of all HSI pixels, respectively, and refer to this division as B(E). For fine tuning the classification part of CNNs over the target sets, we exploited much smaller balanced $\TrainingSet$, $\ValidationSet$, and $\TestSet$ sets with the number of pixels as reported in~\cite{Gao_2018} (the B division). Finally, we simulated MSI with 100, 75, 50, and 25 bands. We report the average accuracy (AA), overall accuracy (OA), and the kappa scores $\kappa=1-\frac{1-p_o}{1-p_e}$, where $p_o$ and $p_e$ are the relative observed agreement, and hypothetical probability of chance agreement, respectively, and $-1\leq \kappa\leq 1$ (the larger $\kappa$ is, the better performance was obtained), elaborated over the unseen test sets, and averaged across 25 executions.

The results for all configurations of 1D-CNN and PT-CNN are gathered in Tables~\ref{tab:summary_our_network}--\ref{tab:summary_sota_network}. For the simulated MSI with 25 bands, some of the models could not be trained due to significant dimensionality reduction performed by the network itself (see the corresponding kernel sizes and strides shown in Fig.~\ref{fig:1d_network} for 1D-CNN, and reported in~\cite{8245897} for PT-CNN). The CNNs which were pre-trained using different source datasets were consistently outperforming those learned over smaller target sets \emph{without} any transfer learning. Therefore, the feature extractors trained over Salinas Valley, Ex(Sa), and Indian Pines, Ex(IP), for 1D-CNN and PT-CNN, respectively, allowed for obtaining the best generalization over the target sets. Increasing the number of CNN building blocks does not bring significant improvement in the classification performance of the underlying models. It shows that even shallower CNNs with notably smaller capacity are able to build appropriate representations of the investigated HSI. Therefore, the most discriminant class features are likely manifested in specific parts of the spectrum, and these features can be automatically elaborated with shallow feature extractors. Finally, we can observe the impact of the dataset split\footnote{We exploit only \emph{spectral} CNNs which operate exclusively on the spectral pixel information during its classification---for such networks, random training-validation-test division \emph{does not} lead to the training-test information leak that makes the classification results over-optimistic and not reliable~\cite{Nalepa2019GRSL}.} on the obtained classification performance of our deep networks---for both 1D-CNN and PT-CNN, the results for B(E) in the full-spectrum scenario were significantly better compared to the B split, where the training sets are much smaller. We report the measures over the unseen $\TestSet$ sets, thus we can conclude that the models trained in the B(E) scheme did not overfit the training data and generalize well. However, this estimated performance was quantified over very limited (and likely not representative) $\TestSet$'s.

In Table~\ref{tab:ranking}, we present the average ranking (according to $\kappa$) of all models trained with and without transfer learning. The dimensionality reduction by using our sliding-window approach allowed us to obtain statistically better performance of both 1D-CNN and PT-CNN when compared with the full HSI in practically all architectural configurations for the simulated MSI with 100, 75, and 50 bands (two-tailed Wilcoxon test at $p<0.005$). The results obtained for the full HSI and the simulated MSI with 25 bands (for 1D-CNN and PT-CNN with 1 building block) and 50 bands (for 1D-CNN and PT-CNN with 3 building blocks) are statistically the same. It shows that the HSI reduction not only does not deteriorate the performance of the models, but can also improve their capabilities. Since the entire spectrum was downsampled, we intrinsically tackled the curse of hyperspectral dimensionality problem. Although simulating MSI from HSI was very beneficial for benchmark scenes, it must be carefully performed for real-life data, as too aggressive HSI reduction can lead to removing parts of the spectrum which convey discriminative information about very \emph{specific} classes, and to making them indistinguishable from other classes with similar spectral profiles.

\begin{table*}[ht!]
\newcommand{\mybackground}{35}
	\scriptsize
	\centering
	\caption{The results obtained for all configurations of the state-of-the-art PT-CNN architecture.}
	\label{tab:summary_sota_network}
	\renewcommand{\tabcolsep}{0.057cm}
	\begin{tabular}{r|r|r|r|r|r|r|r|r|r|r|r|r|r|r|r|r|r|r|r|r|r|r|r|r|r|r|r}
\hline
CNN$\rightarrow$ & \multicolumn{9}{c|}{1 block} & \multicolumn{9}{c|}{2 blocks} & \multicolumn{9}{c}{3 blocks}\\
\hline
 Set$\rightarrow$	& \multicolumn{3}{c|}{Sa}		& \multicolumn{3}{c|}{PU} & \multicolumn{3}{c|}{IP} &
\multicolumn{3}{c|}{Sa}		& \multicolumn{3}{c|}{PU} & \multicolumn{3}{c|}{IP} &
\multicolumn{3}{c|}{Sa}		& \multicolumn{3}{c|}{PU} & \multicolumn{3}{c}{IP} \\	
\hline

&			OA		&AA	 & $\kappa$	&OA 	&AA	& $\kappa$	 &OA		&AA	& $\kappa$	
&			OA		&AA	 & $\kappa$	&OA 	&AA	& $\kappa$	 &OA		&AA	& $\kappa$	
&			OA		&AA	 & $\kappa$	&OA 	&AA	& $\kappa$	 &OA		&AA	& $\kappa$	\\
\hline
Var.$\downarrow$ &\multicolumn{27}{c}{Full spectrum}\\
\hline
B(E)                &93.57&96.68&0.93&96.20&94.97&0.95&77.21&67.93&0.74&94.67&97.27&0.94&96.20&94.96&0.95&86.96&82.84&0.85&94.89&97.50&0.94&96.35&94.99&0.95&86.81&81.48&0.85\\
\hdashline
B                   &\cellcolor{gray!\mybackground}87.07&93.63&\cellcolor{gray!\mybackground}0.86&\cellcolor{gray!\mybackground}83.30&\cellcolor{gray!\mybackground}89.30&0.78&62.33&64.46&0.57&89.62&95.07&0.88&89.29&91.83&0.86&74.40&77.50&0.70&90.34&95.45&0.89&88.46&91.50&0.85&76.26&79.18&0.73\\
\hline
 &\multicolumn{27}{c}{100 bands}\\
\hline
B(E)               &94.51 & 97.39 & 0.94 & 96.38 & 95.11 & 0.93 & 81.88 & 75.61 & 0.81 & 95.45 & 97.83 & 0.95 & 96.56 & 95.41 & 0.95 & 89.11 & 88.15 & 0.90 & 95.32 & 97.76 & 0.95 & 96.43 & 95.22 & 0.96 & 88.67 & 86.05 & 0.90 \\
\hdashline
B                  &88.85 & 94.46 & 0.87 & 88.29 & 91.40 & 0.82 & 65.21 & 68.16 & 0.59 & 90.93 & 95.97 & \cellcolor{red!30}\textbf{0.90} & \cellcolor{red!30}90.40 & \cellcolor{red!30}92.21 & 0.83 & 79.45 & 84.37 & \cellcolor{red!30}0.77 & 90.46 & 95.77 & 0.89 & \cellcolor{red!30}89.32 & \cellcolor{red!30}91.67 & 0.84 & \cellcolor{red!30}79.08 & 82.89 & \cellcolor{red!30}0.76 \\
Ex(Sa)  & --- & --- & --- & \cellcolor{green!40}90.10 & \cellcolor{green!40}92.20 & \cellcolor{green!40}\textbf{0.85} & 68.68 & 72.33 & \cellcolor{green!40}\textbf{0.64} & --- & --- & --- & 89.79 & 92.11 &  \cellcolor{green!40}0.85 &  \cellcolor{green!40}79.91 &  \cellcolor{green!40}\textbf{84.56} & \cellcolor{green!40}0.77 & --- & --- & --- & 89.18 & 91.56 & 0.84 & 78.52 & \cellcolor{green!40}\textbf{83.94} & \cellcolor{green!40}0.76 \\
Ex(PU)  &\cellcolor{green!40}\textbf{89.24} & \cellcolor{green!40}\textbf{94.91} & \cellcolor{green!40}\textbf{0.88} & --- & --- & --- & \cellcolor{green!40}69.48 & \cellcolor{green!40}\textbf{73.11} & \cellcolor{green!40}\textbf{0.64} & 89.29 & 95.15 & 0.89 & --- & --- & --- & 72.69 & 77.46 & 0.68 & 90.18 & 95.46 & 0.89 & --- & --- & --- & 73.23 & 73.23 & 0.68 \\
Ex(IP)  &88.68 & 94.85 & 0.87 & 89.99 & 92.14 & 0.83 & --- & --- & --- & \cellcolor{green!40}\textbf{91.44} & \cellcolor{green!40}\textbf{96.32} & \cellcolor{green!40}\textbf{0.90} & 89.59 & 92.09 & 0.82 & --- & --- & --- & \cellcolor{green!40}\textbf{91.70} & \cellcolor{green!40}\textbf{96.51} & \cellcolor{green!40}\textbf{0.91} & 88.97 & 91.54 & \cellcolor{green!40}0.85 & --- & --- & --- \\
\hline
 &\multicolumn{27}{c}{75 bands}\\
\hline
B(E)                &94.23 & 97.15 & 0.94 & 96.47 & 95.33 & 0.93 & 82.07 & 74.59 & 0.78 & 95.48 & 97.89 & 0.95 & 96.64 & 95.54 & 0.95 & 89.66 & 86.37 & 0.89 & 95.15 & 97.72 & 0.95 & 96.61 & 95.44 & 0.94 & 89.43 & 86.95 & 0.89 \\
\hdashline
B                   &87.97 & 94.03 & \cellcolor{gray!\mybackground}0.86 & 88.61 & 91.70 & 0.84 & 76.35 & 66.63 & 0.59 & 90.55 & 95.86 & \cellcolor{red!30}\textbf{0.90} & \cellcolor{red!30}\textbf{90.42} & \cellcolor{red!30}\textbf{92.63} & 0.84 & \cellcolor{red!30}\textbf{88.89} & \cellcolor{red!30}84.42 & \cellcolor{red!30}\textbf{0.79} & 90.67 & 95.73 & 0.89 & \cellcolor{red!30}\textbf{90.33} & \cellcolor{red!30}\textbf{92.35}& 0.85 & \cellcolor{red!30}\textbf{87.55} & \cellcolor{red!30}83.00 & \cellcolor{red!30}\textbf{0.77} \\
Ex(Sa)   & --- & --- & --- & \cellcolor{green!40}\textbf{90.64} & \cellcolor{green!40}\textbf{92.53} & \cellcolor{green!40}\textbf{0.85} & 78.65 & 70.03 & 0.59 & --- & --- & --- & 90.01 & 92.31 & 0.85 & 87.62 & 84.14 & 0.76 & --- & --- & --- & 89.61 & 91.93 & 0.84 & 87.12 & 81.77 & 0.74 \\
Ex(PU)   &88.14 & 94.40 & \cellcolor{gray!\mybackground}0.86 & --- & --- & --- & \cellcolor{green!40}\textbf{79.65} & \cellcolor{green!40}71.29 & \cellcolor{green!40}0.62 & 89.62 & 95.03 & 0.88 & --- & --- & --- & 80.90 & 73.26 & 0.66 & 90.00 & 95.25 & 0.89 & --- & --- & --- & 82.64 & 76.24 & 0.67 \\
Ex(IP)   &\cellcolor{green!40}88.70 & \cellcolor{green!40}94.70 & \cellcolor{green!40}0.87 & 90.40 & 92.44 & 0.82 & --- & --- & --- & \cellcolor{green!40}91.31 & \cellcolor{green!40}96.24 & \cellcolor{green!40}\textbf{0.90} & 89.86 & 92.23 & \cellcolor{green!40}\textbf{0.87} & --- & --- & --- & \cellcolor{green!40}91.33 & \cellcolor{green!40}96.22 & \cellcolor{green!40}0.90 & 89.32 & 91.71 & \cellcolor{green!40}\textbf{0.86} & --- & --- & --- \\
\hline
 &\multicolumn{27}{c}{50 bands}\\
\hline
B(E)               &94.10 & 97.09 & 0.93 & 96.18 & 95.01 & 0.95 & 80.28 & 71.55 & 0.77 & 95.24 & 97.73 & 0.95 & 96.49 & 95.28 & 0.95 & 87.99 & 85.10 & 0.87 & 94.32 & 97.13 & 0.94 & 96.38 & 95.18 & 0.96 & 87.23 & 81.81 & 0.86 \\
\hdashline
B                  &87.49 & 93.87 & \cellcolor{gray!\mybackground}0.86 & 86.60 & 90.82 & 0.80 & 63.16 & 63.51 & 0.57 & 90.42 & 95.64 & \cellcolor{green!40}0.89 & \cellcolor{red!30}89.47 & \cellcolor{red!30}92.06 & 0.82 & \cellcolor{red!30}76.97 & \cellcolor{red!30}81.96 & \cellcolor{red!30}0.73 & \cellcolor{red!30}90.49 & \cellcolor{red!30}95.79 & \cellcolor{red!30}0.90 & \cellcolor{red!30}88.87 & \cellcolor{red!30}91.53 & \cellcolor{red!30}0.85 & \cellcolor{red!30}75.45 & \cellcolor{red!30}79.64 & \cellcolor{red!30}0.72 \\
Ex(Sa)  & --- & --- & --- & 89.24 & 91.81 & \cellcolor{green!40}\textbf{0.85} & \cellcolor{green!40}67.41 & \cellcolor{green!40}70.34 & \cellcolor{green!40}0.62 & --- & --- & --- & 88.82 & 91.44 & \cellcolor{green!40}0.83 & 74.36 & 79.91 & 0.71 & --- & --- & --- & 83.10 & \cellcolor{gray!\mybackground}88.33 & \cellcolor{gray!\mybackground}0.74 & 69.40 & 75.11 & 0.64 \\
Ex(PU)  &88.20 & 94.29 & 0.87 & --- & --- & --- & 66.53 & 68.41 & 0.61 & \cellcolor{gray!\mybackground}88.62 & \cellcolor{gray!\mybackground}94.52 & \cellcolor{gray!\mybackground}0.88 & --- & --- & --- & \cellcolor{gray!\mybackground}66.58 & \cellcolor{gray!\mybackground}68.68 & \cellcolor{gray!\mybackground}0.61 & \cellcolor{gray!\mybackground}88.95 & \cellcolor{gray!\mybackground}94.77 & \cellcolor{gray!\mybackground}0.88 & --- & --- & --- & \cellcolor{gray!\mybackground}65.13 & \cellcolor{gray!\mybackground}67.80 & \cellcolor{gray!\mybackground}0.58 \\
Ex(IP)  &\cellcolor{green!40}88.25 & \cellcolor{green!40}94.38 & \cellcolor{green!40}0.87 & \cellcolor{green!40}90.13 & \cellcolor{green!40}92.12 & 0.81 & --- & --- & --- & \cellcolor{green!40}90.80 & \cellcolor{green!40}95.91 & \cellcolor{green!40}0.89 & \cellcolor{gray!\mybackground}85.52 & \cellcolor{gray!\mybackground}90.39 & \cellcolor{gray!\mybackground}0.81 & --- & --- & --- & 90.12 & 95.47 & 0.89 & \cellcolor{gray!\mybackground}82.81 & 88.39 & 0.77 & --- & --- & --- \\
\hline
 &\multicolumn{27}{c}{25 bands}\\
\hline
B(E)               &93.09 & 96.50 & 0.92 & 95.90 & 94.55 & 0.94 & 78.08 & 67.60 & 0.75 & --- & --- & --- & --- & --- & --- & --- & --- & --- & --- & --- & --- & --- & --- & --- & --- & --- & ---\\
\hdashline
B                  &87.18 & \cellcolor{gray!\mybackground}93.42 & \cellcolor{gray!\mybackground}0.86 & 84.84 & 89.70 & \cellcolor{gray!\mybackground}0.76 & \cellcolor{gray!\mybackground}61.58 & \cellcolor{gray!\mybackground}62.17 & 0.55 & --- & --- & --- & --- & --- & --- & --- & --- & --- & --- & --- & --- & --- & --- & --- & --- & --- & ---\\
Ex(Sa)  & --- & --- & --- & \cellcolor{green!40}87.81 & \cellcolor{green!40}91.07 & \cellcolor{green!40}0.81 & \cellcolor{green!40}63.52 & \cellcolor{green!40}64.70 & \cellcolor{gray!\mybackground}0.54 & --- & --- & --- & --- & --- & --- & --- & --- & --- & --- & --- & --- & --- & --- & --- & --- & --- & --- \\
Ex(PU)  &87.49 & 93.77 & \cellcolor{gray!\mybackground}0.86 & --- & --- & --- & 63.14 & 63.86 & \cellcolor{green!40}0.56 & --- & --- & --- & --- & --- & --- & --- & --- & --- & --- & --- & --- & --- & --- & --- & --- & --- & --- \\
Ex(IP)  &\cellcolor{green!40}87.95 & \cellcolor{green!40}93.96 & \cellcolor{gray!\mybackground}0.86 & 87.54 & 91.02 & 0.77 & --- & --- & --- & --- & --- & --- & --- & --- & --- & --- & --- & --- & --- & --- & --- & --- & --- & --- & --- & --- & --- \\
\hline
\multicolumn{28}{c}{\scriptsize \textbf{How to read this table:} See the description in Table~\ref{tab:summary_our_network}. }\\
	\end{tabular}
\end{table*}

\begin{table}[ht!]
	\scriptsize
	\centering
	\caption{The average ranking (according to $\kappa$) of all models across all datasets (HSI and simulated MSI). The best ranking is in bold.}
	\label{tab:ranking}
	\renewcommand{\tabcolsep}{0.5cm}
	\begin{tabular}{rrrrrrrrrr}
 \hline
CNN$\downarrow$ & B & Ex(Sa) & Ex(PU) & Ex(IP)\\
\hline
1D-CNN & 1.76 & \textbf{1.41} & 1.59 & 1.45\\
PT-CNN & 1.80 & 1.65 & 1.95 & \textbf{1.55}\\
\hline
	\end{tabular}
\end{table}

To further verify the statistical significance of the obtained results, we executed two-tailed Wilcoxon tests for both CNNs with one, two, and three building blocks, and for all datasets (Table~\ref{tab:wilcoxon}). In the majority of cases, the differences are statistically important ($p<0.005$), thus the deep models in which transfer learning has been applied significantly outperformed those trained over the B target data splits (Table~\ref{tab:ranking}). Also, we can observe the differences between the feature extractors trained over different source HSI datasets---for Pavia University (being the target set), the feature extractors trained over Salinas Valley, Ex(Sa), and Indian Pines, Ex(IP), are statistically the same for both CNNs. Finally, our 1D-CNN outperformed PT-CNN in all scenarios ($p<0.005$). It shows that ensuring the representation invariance with respect to small translation of the input feature maps by the pooling layers is pivotal to get well-generalizing models.

\begin{table}[ht!]
	\scriptsize
	\centering
	\caption{Two-tailed Wilcoxon tests showed that the differences between the investigated CNNs are statistically important in the majority of cases for both architectures with (a)~one, (b)~two, and (c)~three building blocks (the background of statistically important results is green; $p<0.005$).}
	\label{tab:wilcoxon}
	\renewcommand{\tabcolsep}{0.35cm}
	\begin{tabular}{rr|rr|rrrrrr}
 \hline
 & & \multicolumn{2}{c}{\textbf{1D-CNN}} & \multicolumn{2}{c}{\textbf{PT-CNN}}\\
 \hline
 & Set$\rightarrow$ & \multicolumn{4}{c}{\textbf{Salinas Valley}}\\
 \hline
 & &      Ex(PU) & Ex(IP) &      Ex(PU) & Ex(IP)\\
\hline
\multirow{2}{*}{(a)}&
B       & \cellcolor{green!40}$<$0.001	& \cellcolor{green!40}$<$0.001 & \cellcolor{green!40}$<$0.001 &	\cellcolor{green!40}$<$0.001\\
& Ex(PU)  & ---& $>$0.2 &--- & $>$0.1\\
\hdashline
\multirow{2}{*}{(b)}&
B       & \cellcolor{green!40}$<$0.001	& \cellcolor{green!40}$<$0.001 & \cellcolor{green!40}$<$0.001	& \cellcolor{green!40}$<$0.005\\
& Ex(PU)  &--- & \cellcolor{green!40}$<$0.001 & ---& \cellcolor{green!40}$<$0.001\\
\hdashline
\multirow{2}{*}{(c)}&
B       & \cellcolor{green!40}$<$0.005 &	\cellcolor{green!40}$<$0.001 & \cellcolor{green!40}$<$0.001 &	$>$0.05\\
& Ex(PU)  & ---& $>$0.2 & ---& \cellcolor{green!40}$<$0.001\\
\hline
 & Set$\rightarrow$  & \multicolumn{4}{c}{\textbf{Pavia University}}\\
 \hline
 & &      Ex(Sa) & Ex(IP) &      Ex(Sa) & Ex(IP)\\
\hline
\multirow{2}{*}{(a)}&
B       & $>$0.1&	$>$0.1 & \cellcolor{green!40}$<$0.001	& \cellcolor{green!40}$<$0.005\\
& Ex(Sa)  &--- & $>$0.2 & ---& $>$0.2\\
\hdashline
\multirow{2}{*}{(b)}&
B       & \cellcolor{green!40}$<$0.001&	$>$0.05 & $>$0.2	& $>$0.05\\
& Ex(Sa)  &--- & $>$0.05 & ---& $>$0.2\\
\hdashline
\multirow{2}{*}{(c)}&
B       & \cellcolor{green!40}$<$0.01	& $>$0.1 & \cellcolor{green!40}$<$0.05	& \cellcolor{green!40}$<$0.05\\
& Ex(Sa)  &--- & \cellcolor{green!40}$<$0.05 & ---& $>$0.2\\
\hline
 & Set$\rightarrow$  & \multicolumn{4}{c}{\textbf{Indian Pines}}\\
 \hline
 & &      Ex(Sa) & Ex(PU) &      Ex(Sa) & Ex(PU)\\
\hline
\multirow{2}{*}{(a)}&
B       & \cellcolor{green!40}$<$0.001	& \cellcolor{green!40}$<$0.001 & \cellcolor{green!40}$<$0.001	& \cellcolor{green!40}$<$0.001\\
& Ex(Sa)  & --- & \cellcolor{green!40}$<$0.001 &--- & $>$0.2\\
\hdashline
\multirow{2}{*}{(b)}&
B       & \cellcolor{green!40}$<$0.001 &	\cellcolor{green!40}$<$0.01 & $>$0.05	& \cellcolor{green!40}$<$0.001\\
& Ex(Sa)  & --- & $>$0.1 & ---& \cellcolor{green!40}$<$0.001\\
\hdashline
\multirow{2}{*}{(c)}&
B       & \cellcolor{green!40}$<$0.001 &	\cellcolor{green!40}$<$0.001 & \cellcolor{green!40}$<$0.05 &	\cellcolor{green!40}$<$0.001\\
& Ex(Sa)  & --- & $>$0.2 & --- & \cellcolor{green!40}$<$0.001\\
\hline
	\end{tabular}
\end{table}

In Table~\ref{tab:times}, we collect the average training time of all deep feature extractors, the average time of fine tuning the classifiers over the target data, and the average inference time of the trained models for a single example from the unseen test sets $\TestSet$. Although all of the investigated models offered instant inference over all benchmarks, decreasing the spectral dimensionality led to accelerating the inference process. For both CNNs, adding more building blocks, hence increasing the number of trainable CNN parameters, allowed for obtaining faster training convergence, as the capacity of the models are enlarged. Interestingly, training feature extractors in 1D-CNN over the simulated MSI with 75 bands was notably slower than over 100 bands. Since the networks were characterized by the same generalization abilities ($p>0.2$), it indicates that higher-dimensional MSI appeared more challenging to learn from a fairly limited number of training samples. It could be mitigated by either introducing more training examples (i.e.,~generating more ground-truth data points), or---as presented in this letter---by reducing the dimensionality of the training data.

\begin{table*}[ht!]
	\scriptsize
	\centering
	\caption{Average time of feature-extractor training (in s), fine-tuning of the classifiers (s), and the inference of the final models (ms).}
	\label{tab:times}
	\renewcommand{\tabcolsep}{0.4cm}
	\begin{tabular}{r|r|r|r|r|r|r|r|r|r}
\hline
\multicolumn{1}{r|}{CNN$\rightarrow$} & \multicolumn{3}{c|}{1 block} &  \multicolumn{3}{c|}{2 blocks} &  \multicolumn{3}{c}{3 blocks}\\
 \hline
\multicolumn{1}{r|}{Set $\rightarrow$} & \multicolumn{1}{c}{Sa} & \multicolumn{1}{c}{PU} & \multicolumn{1}{c}{IP} & \multicolumn{1}{c}{Sa} & \multicolumn{1}{c}{PU} & \multicolumn{1}{c}{IP} & \multicolumn{1}{c}{Sa} & \multicolumn{1}{c}{PU} & \multicolumn{1}{c}{IP}\\
\hline
Scenario $\downarrow$&\multicolumn{9}{c}{Full spectrum}\\
\hline
1D-CNN, classifier (training) &1562.42&404.74&365.41&1133.83&229.12&221.90&1069.54&268.64&213.91\\
PT-CNN, classifier (training) &651.59&390.51&139.79&483.84&242.10&152.51&536.22&294.72&129.24\\
\hdashline
1D-CNN, classifier (inference) &0.075 & 0.058 & 0.086 & 0.102 & 0.073 & 0.112 & 0.120 & 0.082 & 0.133 \\
PT-CNN, classifier (inference) &0.032 & 0.031 & 0.040 & 0.044 & 0.039 & 0.055 & 0.046 & 0.041 & 0.059 \\
\hline
 &\multicolumn{9}{c}{100 bands}\\
\hline
1D-CNN, extractor &1034.75&274.69&169.55&573.57&180.49&95.77&553.45&221.30&93.56\\
PT-CNN, extractor &1096.81&608.20&223.36&718.79&394.62&196.88&753.72&380.02&190.53\\
\hdashline
1D-CNN, classifier (fine tuning)&169.39&62.59&169.55&134.80&34.15&77.83&118.08&41.27&80.38\\
PT-CNN, classifier (fine tuning)&128.28&66.52&75.68&121.96&45.61&92.91&105.35&48.94&86.01\\
\hdashline
1D-CNN, classifier (inference) &0.057 & 0.057 & 0.066 & 0.072 & 0.072 & 0.085 & 0.082 & 0.082 & 0.096 \\
PT-CNN, classifier (inference) &0.029 & 0.030 & 0.038 & 0.037 & 0.038 & 0.049 & 0.041 & 0.041 & 0.053 \\
\hline
&\multicolumn{9}{c}{75 bands}\\
\hline
1D-CNN, extractor&1347.01&420.51&274.61&998.79&293.05&211.10&638.42&312.06&179.85\\
PT-CNN, extractor&1089.25&575.43&222.08&780.75&394.83&195.71&691.52&448.30&212.69\\
\hdashline
1D-CNN, classifier (fine tuning)&143.41&50.26&94.15&110.17&32.22&68.59&109.64&34.12&64.56\\
PT-CNN, classifier (fine tuning)&124.44&70.17&90.18&127.59&52.28&195.71&100.21&55.21&88.21\\
\hdashline
1D-CNN, classifier (inference) &0.050 & 0.050 & 0.061 & 0.061 & 0.063 & 0.073 & 0.071 & 0.072 & 0.084 \\
PT-CNN, classifier (inference) &0.030 & 0.030 & 0.039 & 0.038 & 0.039 & 0.051 & 0.041 & 0.040 & 0.056 \\
\hline
&\multicolumn{9}{c}{50 bands}\\
\hline
1D-CNN, extractor&1071.76&366.89&217.80&686.72&250.31&180.00&564.22&274.33&144.77\\
PT-CNN, extractor&1072.14&636.39&228.29&756.48&443.57&207.35&589.82&538.99&196.65\\
\hdashline
1D-CNN, classifier (fine tuning)&116.23&49.21&77.35&91.71&32.78&60.51&80.17&37.34&59.89\\
PT-CNN, classifier (fine tuning)&125.65&72.40&88.44&119.24&51.11&92.58&117.40&55.90&91.65\\
\hdashline
1D-CNN, classifier (inference) &0.047 & 0.046 & 0.054 & 0.051 & 0.059 & 0.069 & 0.060 & 0.066 & 0.081 \\
PT-CNN, classifier (inference) &0.029 & 0.030 & 0.037 & 0.038 & 0.039 & 0.048 & 0.041 & 0.040 & 0.051 \\
\hline
&\multicolumn{9}{c}{25 bands}\\
\hline
1D-CNN, extractor&749.85&397.06&166.28&470.53&250.17&117.43&---&---&---\\
PT-CNN, extractor&1083.01&756.97&222.39&---&---&---&---&---&---\\
\hdashline
1D-CNN, classifier (fine tuning)&88.59&41.18&60.51&66.96&31.97&48.27&---&---&---\\
PT-CNN, classifier (fine tuning)&127.05&74.16&103.64&---&---&---&---&---&---\\
\hdashline
1D-CNN, classifier (inference) &0.037 & 0.039 & 0.046 & 0.051 & 0.052 & 0.060 &---&---&--- \\
PT-CNN, classifier (inference) &0.029&0.029&0.038&---&---&---&---&---&---\\
\hline
	\end{tabular}
\end{table*}

\section{Conclusion}\label{sec:conclusions}

In this letter, we tackled the problem of limited ground-truth hyperspectral data in the context of supervised hyperspectral image segmentation. We utilize transfer learning, train the deep models over a source set, and apply the learned feature extractors to the target data after fine tuning the classification part of a CNN. We made our method applicable to \emph{any} input HSI by incorporating effective dimensionality reduction, and simulated a constant number of bands for source and target sets. Our multi-faceted experimental study showed that the models trained with transfer learning significantly (in the statistical sense) outperformed the other CNNs, and that our dimensionality reduction not only does not adversely affect the performance of the models, but improves their generalization. It brings new possibilities for on-board deep learning-powered Earth observation use cases, where transferring full HSI data is extremely costly, and the lack of ground-truth is an important real-life obstacle in deploying such learners in the wild.

\ifCLASSOPTIONcaptionsoff
  \newpage
\fi

\bibliographystyle{ieeetran}
\bibliography{IEEEabrv,ref_all}

\end{document}